\documentclass[10pt,twocolumn,letterpaper]{article}

\usepackage{cvpr}
\usepackage{times}
\usepackage{epsfig}
\usepackage{graphicx}
\usepackage{amsmath}
\usepackage{amssymb}

\usepackage{booktabs,subcaption,amsfonts,dcolumn,stfloats}
\usepackage[ruled, noline]{algorithm2e}

\usepackage[pagebackref=true,breaklinks=true,letterpaper=true,colorlinks,bookmarks=false]{hyperref}

\cvprfinalcopy 

\begin{document}

\title{Blockwisely Supervised Neural Architecture Search with Knowledge Distillation}

\author{Changlin Li$^{1,2}$\thanks{Changlin Li and Jiefeng Peng contribute equally and share first-authorship. This work was done when Changlin Li worked as an intern.}, \quad Jiefeng Peng$^{1}$\footnotemark[1],\\ \quad Liuchun Yuan$^{1,3}$, \quad Guangrun Wang$^{1,3}$\thanks{Corresponding Author is Guangrun Wang.}, \quad Xiaodan Liang$^{1,3}$, \quad Liang Lin$^{1,3}$, \quad Xiaojun Chang$^{2}$\\
\small$^1$DarkMatter AI Research \quad \small$^2$Monash University \quad \small$^3$Sun Yat-sen University\\
{\tt\small \{jiefengpeng,ylc0003,xdliang328\}@gmail.cn,} \tt\small wanggrun@mail2.sysu.edu.cn,\\ \tt\small linliang@ieee.org, 
{\tt\small \{changlin.li,xiaojun.chang\}@monash.edu}
}

\maketitle

\begin{abstract}
\vspace{-10pt}
Neural Architecture Search (NAS), aiming at automatically designing network architectures by machines, is hoped and expected to bring about a new revolution in machine learning. Despite these high expectation, the effectiveness and efficiency of existing NAS solutions are unclear, with some recent works going so far as to suggest that many existing NAS solutions are no better than random architecture selection. The inefficiency of NAS solutions may be attributed to inaccurate architecture evaluation. Specifically, to speed up NAS, recent works have proposed under-training different candidate architectures in a large search space concurrently by using shared network parameters; however, this has resulted in incorrect architecture ratings and furthered the ineffectiveness of NAS.

In this work, we propose to modularize the large search space of NAS into blocks to ensure that the potential candidate architectures are fully trained; this reduces the representation shift caused by the shared parameters and leads to the correct rating of the candidates. Thanks to the block-wise search, we can also evaluate all of the candidate architectures within a block. Moreover, we find that the knowledge of a network model lies not only in the network parameters but also in the network architecture. Therefore, we propose to distill the neural architecture (DNA) knowledge from a teacher model as the supervision to guide our block-wise architecture search, which significantly improves the effectiveness of NAS. Remarkably, the capacity of our searched architecture has exceeded the teacher model, demonstrating the practicability and scalability of our method.
Finally, our method achieves a state-of-the-art 78.4\% top-1 accuracy on ImageNet in a mobile setting, which is about a 2.1\% gain over EfficientNet-B0. All of our searched models along with the evaluation code are available at \url{https://github.com/changlin31/DNA}.
\end{abstract}

\vspace{-16pt}
\section{Introduction}\label{sec:intro}
\vspace{-6pt}
Due to the importance of automatically designing machine learning algorithms using machines, interest in the prospect of Automated Machine Learning (AutoML) has been a growing recently. Neural architecture search (NAS), as an essential task of AutoML, is hoped and expected to reduce the effort required to be expended by human experts in network architecture design. Research into NAS has been accelerated in the past two years by the industry, and a number of solutions have been proposed.
However, the effectiveness and efficiency of existing NAS solutions are unclear. Typically, \cite{sciuto2019evaluating} and \cite{yang2019evaluation} even suggest that many existing solutions to NAS are no better than or struggle to outperform random architecture selection. Hence, the question of how to efficiently solve a NAS problem remains an active and unsolved research topic.

\begin{figure}[t]
\vspace{-20pt}
\setlength{\abovecaptionskip}{0cm}
\setlength{\belowcaptionskip}{-0.7cm}
\centering
\includegraphics[width=0.4\textwidth]{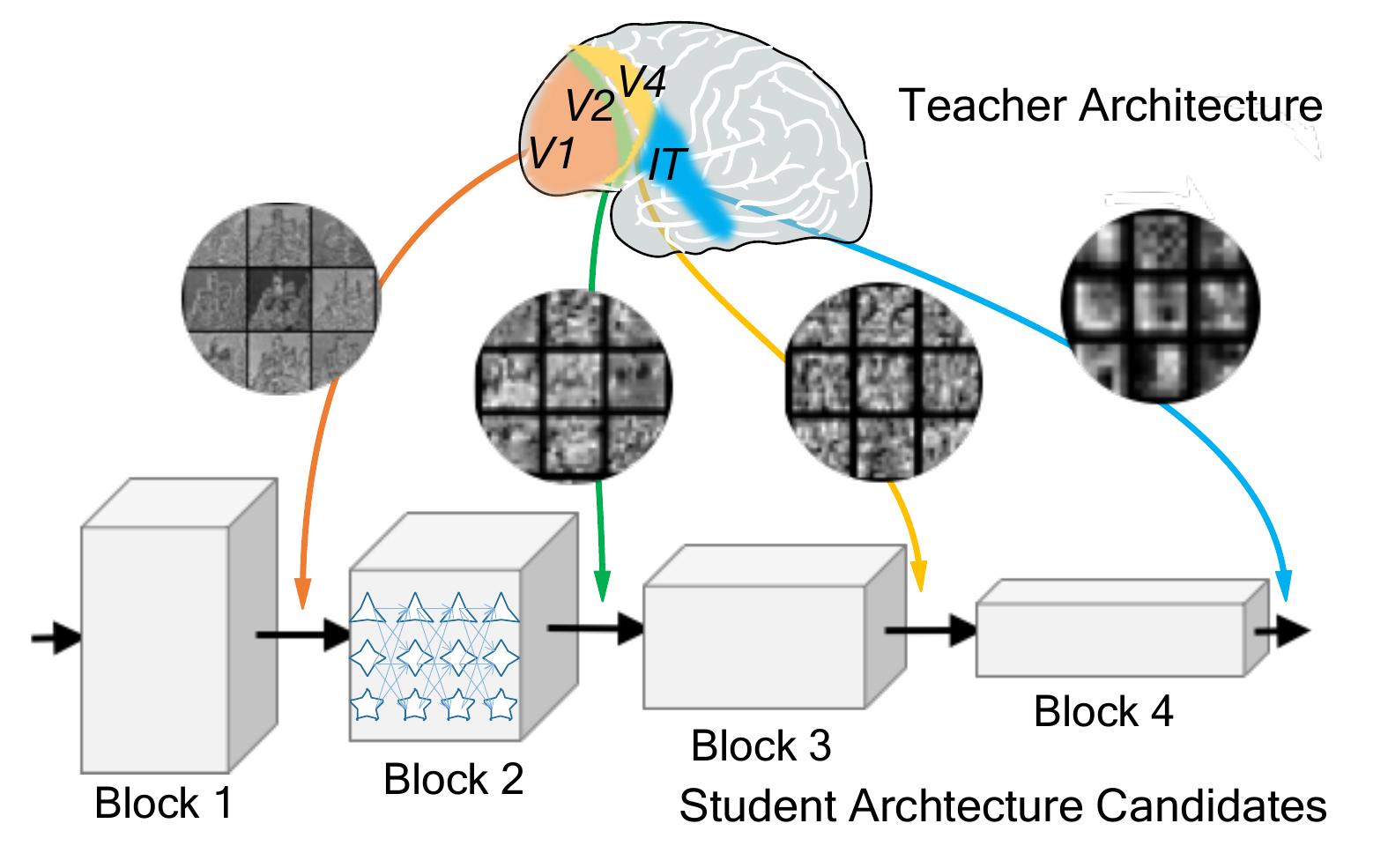}
\caption{\small{We consider a network architecture has several blocks, conceptualized as analogous to the ventral visual blocks $\mathcal{V}1$, $\mathcal{V}2$, $\mathcal{V}4$, and $\mathcal{IT}$ \cite{shipp1985segregation}. Then, we search for the candidate architectures (denoted by different shapes and paths) block-wisely guided by the architecture knowledge distilled from a teacher model.}}\label{fig:intro}
\end{figure}

The most mathematically accurate solution to NAS is to train each of the candidate architectures within the search space from scratch to convergence and compare their performance; however, this is impractical due to the astonishingly high cost. A suboptimal solution is to train only the architectures in a search \emph{sub-space} using advanced search strategies like Reinforcement Learning (RL) or Evolutionary Algorithms (EA); although this is still time-consuming, as training even \emph{one} architecture costs a long time (e.g., more than 10 GPU days for a ResNet on ImageNet). To speed up NAS, recent works have proposed that rather than training each of the candidates fully from scratch to convergence, different candidates should be trained concurrently by using shared network parameters. Subsequently, the ratings of different candidate architectures can be determined by evaluating their performance based on these undertrained shared network parameters. However, several questions remain: does the evaluation based on the undertrained network parameters correctly rank the candidate models? Can the architecture that achieves the highest accuracy defend its top ranking when trained from scratch to convergence?
\cite{chu2019fairnas} and \cite{li2019improving} have suggested that when the search space is small and all the candidates fully and fairly trained, the answer to the above questions is guaranteed to be ``\emph{yes}''. Unfortunately, it is not recommended to narrow down the search space, as a small search space will lead to a very narrow accuracy range, making the search meaningless.

To address the above-mentioned issues, we propose a new solution to NAS where the search space is large, while the potential candidate architectures can be fully and fairly trained. We consider a network architecture that has several blocks, conceptualized as analogous to the ventral visual blocks $\mathcal{V}1$, $\mathcal{V}2$, $\mathcal{V}4$, and $\mathcal{IT}$ \cite{shipp1985segregation} (see Fig. \ref{fig:intro}). We then train each block of the candidate architectures separately. As guaranteed by the mathematical principle, the number of candidate architectures in a block reduces exponentially compared to the the number of candidates in the whole search space. Hence, the architecture candidates can be fully and fairly trained, while the representation shift caused by the shared parameters is reduced, leading to the correct candidate ratings. The correct and visiting-all evaluation improves the effectiveness of NAS. Moreover, thanks to the modest amount of the candidates in a block, we can even search for the depth of a block, which further improves the performance of NAS.

Moreover, lack of supervision for the hidden block creates a technical barrier in our greedy block-wise search of network architecture. To deal with this problem, we propose a novel knowledge distillation method, called \textbf{DNA}, that distills the neural architecture from an existing architecture. As Fig. \ref{fig:intro} shows, we find that different blocks of an existing architecture have different knowledge in extracting different patterns of an image. For example, the lowest block acts like the $\mathcal{V}1$ of the ventral visual area, which extracts low-level features of an image, while the upper block acts like the $\mathcal{IT}$ area, which extracts high-level features. We also find that the knowledge not only lies, as the literature suggests, in the network parameters, but also in the network architecture. Hence, we use the block-wise representation of existing models to supervise our architecture search. Note that the capacity of our searched architectures is not bounded by the capacity of the supervising model. We have searched a number of architectures that have fewer parameters but significantly outperforms the supervising model, demonstrating the practicability and scalability of our DNA method.

Furthermore, inspired by the remarkable success of the transformers (e.g., BERT \cite{devlin2018bert} and \cite{vaswani2017attention}) in natural language domain that discard the inefficient sequential training of RNN, we propose to parallelize the block-wise search in an analogous way. Specifically, for each block, we use the output of the previous block of the supervising model as the input for each of our blocks. Thus, the search can be sped up in a parallel way.

Overall, our contributions are three-fold:\begin{itemize}
\vspace{-6pt}
\item{} We propose to modularize the large search space of NAS into blocks, ensuring that the potential candidate architectures are fairly trained, and the representation shift caused by the shared parameters is reduced, which leads to correct ratings of the candidates. The correct and visiting-all ratings improve the effectiveness of NAS. Novelly, we also search for the depth of the architecture with the help of our block-wise search.
\vspace{-16pt}
\item{} We find that the knowledge of a network model lies not only, as the literature suggests, in the network parameters, but also in the network architecture. Therefore, we use the architecture knowledge distilled from a teacher model to guide our block-wise architecture search. Remarkably, the capacity of our searched architecture has exceeded the teacher model, proving the practicability and scalability of our proposed DNA.
\vspace{-6pt}
\item{} Strong empirical results are obtained on ImageNet and CIFAR10. Typically, our DNA with 6.4M parameters obtains a 78.4\% top-1 accuracy on ImageNet, which is about 2.1\% higher than the result obtained by EfficientNet-B0 with a similar parameter number. To the best of our knowledge, this is the state-of-the-art model in a mobile setting.
\end{itemize}

\vspace{-12pt}
\section{Related Work}\label{sec:related work}
\vspace{-4pt}
\noindent\textbf{Neural Architecture Search (NAS).} NAS is hoped to replace the effort of human experts in network architecture design by machines. Early works \cite{zoph2017neural, baker2017designing, zhong2018practical, chen2019renas, negrinho2018deeparchitect} adopt an agent (e.g., an RNN or an EA method) to sample an architecture and get its performance through a complete training procedure. This type of NAS is computationally expensive and difficult to deploy on large-datasets.

More recent studies \cite{cai2019proxylessnas, liu2019darts, dong2019searching, AkimotoICML2019, brock2018smash} encode the entire search space as a weight sharing supernet. Gradient-based approches\cite{liu2019darts, cai2019proxylessnas, wu2019fbnet} jointly optimize the weight of the supernet and the architecture choosing factors by gradient descent. However, optimizing these choosing factors brings inevitable bias between sub-models. Since the sub-model performing poor in the beginning will get trained less and easily stay behind others, these methods depend heavily on the initial states, making it difficult to reach the best architecture. One-shot approaches \cite{guo2019single, chu2019fairnas, brock2018smash, bender2018understanding} ensure fairness among all sub-models. After training the supernet via path dropout or path sampling, sub-models are sampled and evaluated with the weights inherited from the supernet. However, as identified in \cite{bender2018understanding, chu2019fairnas, li2019improving}, there is a gap between the accuracy of the proxy sub-model with shared weights and the retrained stand-alone one. This gap narrows as the amount of weight sharing sub-models decrease \cite{chu2019fairnas, li2019improving}.\\

\vspace{-8pt}
\noindent\textbf{Knowledge Distillation.} Knowledge distillation is a classical method of model compression, which aims at transferring knowledge from a trained \emph{teacher} network to a smaller and faster \emph{student} model. Existing works on knowledge distillation can be roughly classified into two categories. The first category is to use soft-labels generated by the teacher to teach the student, which is first proposed by \cite{ba2014deep}. Later, Hinton \etal \cite{hinton2014distilling} redefined knowledge distillation as training a shallower network to approach the teacher's output after the softmax layer. However, when the teacher model gets deeper, learning the soft-labels alone is insufficient. To address this problem, the second category of knowledge distillation proposes to employ the internal representation of the teacher to guide the training of the student \cite{romero2015fitnets, zhang2017knowledge, yim2017gift, wang2018progressive, passalis2018learning}. \cite{yim2017gift} proposed a distillation method to train a student network to mimic the teacher's behavior in multiple hidden layers jointly. \cite{wang2018progressive} proposed a progressive block-wise distillation to learn from several of the teacher's intermediate feature maps, which eases the difficulty of joint optimization but increases the gap between the student and the teacher model during the progressive distillation. All existing works assume that the knowledge of a network model lies in the network parameter, while we find that the knowledge also lies in the network architecture. Moreover, in contrast to \cite{wang2018progressive} , we proposed a parallelized distillation procedure to reduce both the gap and the time consumption.
\section{Methodology}

\vspace{-4pt}
\noindent
We begin with the inaccurate evaluation problem of existing NAS, based on which we define our block-wise search.
\vspace{-3pt}
\subsection{Challenge of NAS and our Block-wise Search}\label{defination}

\vspace{-4pt}
Let $\alpha \in \mathcal{A}$ and $\omega_{\alpha}$ denote the network architecture and the network parameters, respectively, where $\mathcal{A}$ is the architecture search space. A NAS problem is to find the optimal pair ($\alpha^*$, $\omega^*_\alpha$) such that the model performance is maximized. Solving a NAS problem often consists of two iterative steps, i.e., search and evaluation. A search step is to select an appropriate architecture for evaluation, while an evaluation step is to rate the architecture selected by the search step. The evaluation step is of most importance in the solution to NAS because an inaccurate evaluation leads to the ineffectiveness of NAS, and a slow evaluation results in the inefficiency of NAS.\\

\vspace{-8pt}
\noindent\textbf{Inaccurate Evaluation in NAS.} The most mathematically accurate evaluation for a candidate architecture is to train it from scratch to convergence and test its performance, which, however, is impractical due to the awesome cost. For example, it may cost more than 10 GPU days to train a ResNet on ImageNet. To speed up the evaluation, recent works \cite{bender2018understanding, liu2019darts, cai2019proxylessnas, guo2019single, wu2019fbnet, li2019improving} propose not to train each of the candidates fully from scratch to convergence, but to train different candidates concurrently by using shared network parameters. Specifically, they formulate the search space $\mathcal{A}$ into an over-parameterized supernet such that each of the candidate architecture $\alpha$ is a sub-net of the supernet. Let $\mathcal{W}$ denote the network parameters of the supernet. The learning of the supernet is as follows:
\begin{small}\begin{equation}\label{eq:loss-supernet}
\mathcal{W}^* = \min_\mathcal{W} \mathcal{L}_{\text{train}}(\mathcal{W}, \mathcal{A}; \mathbf{X}, \mathbf{Y}),
\end{equation}\end{small}where $\mathbf{X}$ and $\mathbf{Y}$ denote the input data and the ground truth labels, respectively. Here, $\mathcal{L}_{\text{train}}$ denotes the training loss. Then, the ratings of different candidate architectures are determined by evaluating their performance based on these shared network parameters, $\mathcal{W}^*$.

However, as analyzed in Section \ref{sec:intro}, the optimal network parameter $\mathcal{W}^*$ does not necessarily indicate the optimal network parameters $\omega^*$ for the sub-nets (i.e., the candidate architectures) because the sub-nets are not fairly and fully trained. The evaluation based on $\mathcal{W}^*$ does not correctly rank the candidate models because the search space is usually large (e.g., $>1e^{15}$). The inaccurate evaluation has led to the ineffectiveness of the existing NAS.\\

\vspace{-8pt}
\noindent\textbf{Block-wise NAS.} \cite{chu2019fairnas} and \cite{li2019improving} have suggested that when the search space is small, and all the candidates are fully and fairly trained, the evaluation could be accurate. To improve the accuracy of the evaluation, we divide the supernet into blocks of smaller sub-space. Specifically, Let $\mathcal{N}$ denote the supernet. We divide $\mathcal{N}$ into $N$ blocks by the depth of the supernet and have:
\begin{small}\begin{equation}
  \mathcal{N}= \mathcal{N}_N \dots \mathcal{N}_{i+1} \circ \mathcal{N}_{i} \dots \circ \mathcal{N}_1,
\end{equation}
\end{small}where $\mathcal{N}_{i+1} \circ \mathcal{N}_{i}$ denotes that the ($i+1$)-th block is originally connected to the $i$-th block in the supernet. Then we learn each block of the supernet separately using:
\begin{equation}\label{eq:loss-block}
\mathcal{W}_i^* = \min_{\mathcal{W}_i} \mathcal{L}_{\text{train}}(\mathcal{W}_i, \mathcal{A}_i; \mathbf{X}, \mathbf{Y}), ~~~~i = 1,2\cdots,N,
\end{equation}where $\mathcal{A}_i$ denote the search space in the $i$-th block.

\begin{figure*}
\vspace{-11pt}
\setlength{\abovecaptionskip}{0cm}
\setlength{\belowcaptionskip}{-0.2cm}
\begin{center}
\includegraphics[width=1\linewidth]{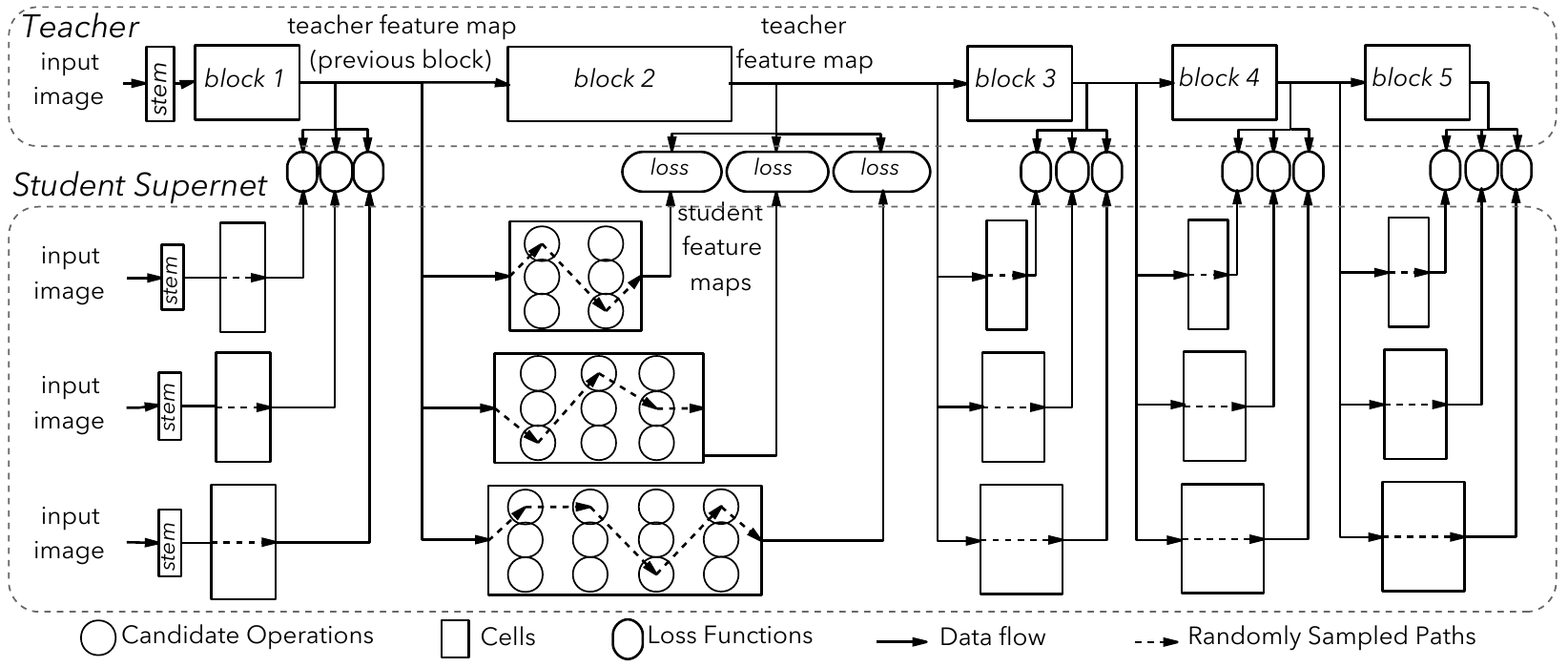}
\end{center}
\vspace{-11pt}
  \caption{\small{Illustration of our DNA. The teacher's previous feature map is used as input for both teacher and student block. Each cell of the supernet is trained independently to mimic the behavior of the corresponding teacher block by minimizing the l2-distance between their output feature maps. The dotted lines indicate randomly sampled paths in a cell.}}\label{fig:training}
  \vspace{-11pt}
\end{figure*}

Are the candidate architectures in each block fully trained? How large is the search space in a block? Let $d$ denote the depth of the $i$-th block and $C$ denote the number of the candidate operations in each layer. Then the size of the search space of the $i$-th block is $C^{d_i},\forall i \in [1, N]$; the size of the search space $\mathcal{A}$ is \begin{small}$\prod\limits_{i=0}^N C^{d_i}$\end{small}. This indicates a exponential drop in the size of the search space: \begin{equation}\label{eq:space-size}
\text{Drop rate} = {C^{d_i}}/{(\prod\limits_{i=0}^N C^{d_i})}.
\end{equation}In our experiment, the search space in a block reduces significantly (e.g., $\text{Drop rate}\approx 1/(1e^{\frac{15}{N}}$)), ensuring each candidate architecture $\alpha_i \in \mathcal{A}_i$ to be optimized sufficiently. Finally, the architecture is searched across the different blocks in the whole search space $\mathcal{A}$:
\begin{small}\begin{equation}
  \alpha^* = \mathop{\arg\min}_{\alpha \in \mathcal{A}} \sum_{i=1}^{N} \lambda_i \mathcal{L}_{\text{val}}(\mathcal{W}_i^* (\alpha_i), \alpha_i; \mathbf{X}, \mathbf{Y}),
\end{equation}\end{small}where $\lambda_i$ represents the loss weights. Here, $\mathcal{W}_i^* (\alpha_i)$ denotes the learned shared network parameters of the sub-net $\alpha_i$ and the supernet. Note that different from the learning of the supernet, we use the validation set to evaluate the performance of the candidate architectures.

\subsection{Block-wise Supervision with Distilled Architecture Knowledge}\label{sec:supernetdistill}

\vspace{-2pt}
Although we motivate well in Section \ref{defination}, a technical barrier in our block-wise NAS is that we lack of internal ground truth in Eqn. \eqref{eq:loss-block}. Fortunately, we find that different blocks of an existing architecture have different knowledge\footnote{The definition of knowledge is a matter of ongoing debate among philosophers. In this work, we specially define KNOWLEDGE as follows. \textbf{Knowledge} is the skill to recognize some patterns; \textbf{Parameter Knowledge} is the skill of using appropriate network parameter to recognize some patterns. \textbf{Architecture Knowledge} is the skill of using appropriate network structrue to recognize some patterns.} in extracting different patterns of an image. We also find that the knowledge not only lies, as the literature suggests, in the network parameters, but also in the network architecture. Hence, we use the block-wise representation of existing models to supervise our architecture search. Let $\mathcal{Y}_i$ be the output feature maps of the $i$-th block of the supervising model (i.e., teacher model) and $\hat{\mathcal{Y}}_i(\mathcal{X})$ be the output feature maps of the $i$-th block of the supernet. We take L2 norm as the cost function. The loss function in Eqn. \eqref{eq:loss-block} can be written as:\begin{small}\begin{equation}\label{eq:distill-loss}
\mathcal{L}_{\text{train}}(\mathcal{W}_i, \mathcal{A}_i; \mathbf{X}, \mathcal{Y}_i) = \frac{1}{K} \left\| \mathcal{Y}_i - \hat{\mathcal{Y}}_i(\mathcal{X}) \right\|_2^2,
\end{equation}
\end{small}where $K$ denotes the numbers of the neurons in $\mathcal{Y}$.

Moreover, inspired by the remarkable success of the transformers (e.g., BERT \cite{devlin2018bert} and \cite{vaswani2017attention}) in natural language domain that discards the inefficient sequential training of RNN, we propose to parallelize the block-wise search in an analogous way. Specifically, for each block, we use the output $\mathcal{Y}_{i-1}$ of the ($i-1$)-th block of the teacher model as the input of the $i$-th block of the supernet. Thus, the search can be sped up in a parallel way. Eqn. \eqref{eq:distill-loss} can be written as:\begin{small}\begin{equation}\label{eq:distill-loss-re}
\mathcal{L}_{\text{train}}(\mathcal{W}_i, \mathcal{A}_i; \mathcal{Y}_{i-1}, \mathcal{Y}_i) = \frac{1}{K} \left\| \mathcal{Y}_i - \hat{\mathcal{Y}}_i(\mathcal{X}) \right\|_2^2,
\end{equation}
\end{small}

\vspace{-12pt}
Note that the capacity of our searched architectures is not bounded by the capacity of the supervising model, e.g., we have searched a number of architectures that have fewer parameters but significantly beats the supervising model. By scaling our architecture to the same model size as the supervising architecture, a more remarkable gain is further obtained, demonstrating the practicability and scalability of our DNA. Fig.\ref{fig:training} shows a pipeline of our block-wise supervision with knowledge distillation.

\vspace{-6pt}
\subsection{Automatic Computation Allocation with Channel and Layer Variability}\label{sect_scal}

\vspace{-2pt}
Automatically allocating model complexity of each block is especially vital when performing block-wise NAS under a certain constraint. To better imitate the teacher, the model complexity of each block may need to be allocated according to the learning difficulty of the corresponding teacher block adaptively. With the input image size and the stride of each block fixed, generally, the computation allocation is only related to the width and depth of each block, which are burdensome to search in a weight sharing supernet. Both the width and depth are usually pre-defined when designing the supernet for a one-shot NAS method. Most previous works include \emph{identity} as a candidate operation to increase supernet scalability \cite{bender2018understanding, liu2019darts, cai2019proxylessnas, wu2019fbnet, li2019improving}. However, as pointed out in \cite{chu2019scarletnas}, adding identity as a candidate operation can lead to convergence difficulty of the supernet, as well as an unfair comparison of sub-models. In addition, adding identity as a candidate operation may lead to a detrimental and unnecessary increase in the possible sequence of operations. For example, a sequence of operation \{conv, identity, conv\} is equivalent to \{conv, conv, identity\}. This unnecessary increase of search space results in a drop of the supernet stability and fairness. Besides, \cite{liang2020computation} searches for the layer number with fixed operations for the first step, and subsequently searched for three operations with a fixed layer number. However, the choice of operations is not independent from the layer number of each block. To search for more candidate operations by this two-step method could lead to a bigger gap from the real target.

Thanks to our block-wise search, we can train several \emph{cells} with different channel numbers or layer numbers independently in each stage to ensure channel and layer variability without the interference of identity operation, As shown in Figure \ref{fig:training}, in each training step, the teacher's previous feature map is first fed to several cells (as suggested by the solid line), and one of the candidate operations of each layer in the cell is randomly chosen to form a path (as suggested by the dotted line). The weight of the supernet is optimized by minimizing the MSE loss with the teacher's feature map.

\vspace{-2pt}

\subsection{Searching for Best Student Under Constraint}\label{sec:search}

\vspace{-4pt}
Our typical supernet contains about $10^{17}$ sub-models, which stops us from evaluating all of them. In previous one-shot NAS methods, random sampling, evolutionary algorithms and reinforcement learning have been used to sample sub-models from the trained supernet for further evaluation. In most recent work \cite{liang2020computation, li2019improving}, a greedy search algorithm is used to progressively shrink the search space by selecting the top-performing partial models layer by layer. Considering our block-wise distillation, we propose a novel method to estimate the performance of all sub-models according to their block-wise performance and subtly traverse all the sub-models to select the top-performing ones under certain constraints.\\

\vspace{-8pt}
\begin{algorithm}
\footnotesize
\caption{Feature sharing evaluation}\label{alg:eval}
\Indmm

\SetKw{Def}{define}
\SetKw{Out}{output}
\SetKwFunction{dfsforward}{DFS-Forward}
 \KwIn{Teacher's previous feature map $G_{prev}$, Teacher's current feature map $G_{curr}$, Root of the cell $Cell$, loss function $loss$}
 \KwOut{List of evaluation loss $L$}
\Indp
  \BlankLine
  \Def \dfsforward{$N$, $X$}:\\
\Indp
  $Y = N(X)$\;
  \eIf{$N$ has no $child$}{
    $append(L, loss(Y, G_{curr}))$\;}
  {
    \For{$C$ in $N.child$}{
      \dfsforward{$C$, $Y$}\;}}
\Indm
\BlankLine
\dfsforward{$Cell$, $G_{prev}$}\;
\Out{$L$}\;
\end{algorithm}
\setlength{\floatsep}{0.1cm}
\setlength{\textfloatsep}{0.1cm}
\begin{algorithm}
\footnotesize
\caption{Traversal search}\label{alg:search}
\Indmm
\KwIn{Block index $B$, the teacher's current feature map $G$, constrain $C$, model pool list $Pool$}
\KwOut{best model $M$}
\SetKw{Def}{define}
\SetKw{Out}{output}
\SetKwFunction{traverseblock}{SearchBlock}
\Indpp
  \Def \traverseblock{$B, size_{prev}, loss_{prev}$}:\\
\Indp
  \For{$i < length(Pool[B])$}{
  $size \leftarrow size_{prev} + size[i]$\;
  \If{$size > C$}{continue\;}
  $loss \leftarrow loss_{prev} + loss[i]$\;
  \eIf{B is last block}
  { 
    \If{$loss \leq loss_{best}$}
      {$loss_{best} \leftarrow loss$\;
      $M \leftarrow$ index of each block
    }
    break\;}
  {\traverseblock{$B+1, size, loss$}\;}
}
\Indm
\BlankLine
\traverseblock{0}\;
\Out{$M$}\;
\end{algorithm}

\noindent\textbf{Evaluation.} In our method, we aim to imitate the behavior of the teacher in every block. Thus, we estimate the learning ability of a student sub-model by its evaluation loss in each block. Our block-wise search make it possible to evaluate all the partial models (about $10^4$ in each cell). To accelerate this process, we forward-propagate a batch of input node by node in a manner similar to deep first search, with intermediate output of each node saved and reused by subsequent nodes to avoid recalculating it from the beginning. The feature sharing evaluation algorithm is outlined in Algorithm \ref{alg:eval}. By evaluating all cells in a block of the supernet, we can get the evaluation loss of all possible paths in one block. We can easily sort this list with about $10^4$ elements in a few seconds with a single CPU. After this, we can select the top-1 partial model from every block to assemble a best student. However, we still need to find efficient models under different constraints to meet the needs of real-life applications.\\

\vspace{-8pt}
\noindent\textbf{Searching.} After performing evaluation and sorting, the partial model rankings of each stage are used to find the best model under a certain constraint. To automatically allocate computational costs to each block, we need to make sure that the evaluation criteria are fair for each block. We notice that MSE loss is related to the size of the feature map and the variance of the teacher's feature map. To avoid any possible impact of this, a fair evaluation criterion, called relative $l1$ loss, is defined as:
\begin{equation}
  \mathcal{L}_R(x, y) = \frac{{||x-y||}_{1}}{\sigma(y)},
  \vspace{-4pt}
\end{equation}
where $\sigma(\cdot)$ means standard deviation among all elements. All the $\mathcal{L}_R$ in each block of a sub-model is added up to estimate the ability to learn from the teacher. However, it is unnecessarily time-consuming to calculate the complexity and add up the loss for all $10^{17}$ candidate models. With ranked partial models in each block, a time-saving search algorithm (Alg. \ref{alg:search}) is proposed to visit all possible models subtly. Note that we get the complexity of each candidate operation by a precalculated lookup table to save the time. The testing of next block is skipped if current partial model combining with the smallest partial model in the following blocks already exceed the constraint. Moreover, it returns to the previous block after finding a model satisfying the constraint, to prevent testing of subsequent models with lower rank in current block.
\vspace{-8pt}

\section{Experiments}

\vspace{-4pt}
\subsection{Setups}\label{sec:setup}
\vspace{-4pt}
\noindent\textbf{Choice of dataset and teacher model.} We evaluated our method on ImageNet, a large-scale classification dataset that has been used to evaluate various NAS methods. We randomly select 50 images from each class of the original training set to form a 50000-image validation set for search procedure and use the remainder as training set. Note that all of our results are tested on the original validation set.

\begin{table}
\setlength{\abovecaptionskip}{0.cm}
\setlength{\belowcaptionskip}{0cm}
\begin{center}
\caption{Our supernet design. ``$l\#$'' and ``ch$\#$'' means layer and channel number of each cell.}
\label{tab:design}

    \begin{tabular}[t]{p{0.8cm}|p{0.3cm}p{0.5cm}|p{0.3cm}p{0.5cm}|p{0.3cm}p{0.5cm}|p{0.3cm}p{0.5cm}}
    \hline
    model&\multicolumn{2}{c|}{teacher}&\multicolumn{6}{c}{student supernet}\cr\hline
    block & $l\#$ & ch$\#$ & $l\#$ & ch$\#$ & $l\#$ & ch$\#$ & $l\#$ & ch$\#$ \\
    \hline
    1 & 7 & 48 & 2 & 24 & 3 & 24 & 2 & 32\\
    2 & 7 & 80 & 2 & 40 & 3 & 40 & 4 & 40\\
    3 & 10 & 160 & 2 & 80 & 3 & 80 & 4 & 80\\
    4 & 10 & 224 & 3 & 112 & 4 & 112 & 4 & 96\\
    5 & 13 & 384 & 4 & 192 & 5 & 192 & 5 & 160\\
    6 & 4 & 640 & 1 & 320 & - & - & - & -\\
    \hline
    \end{tabular}
\vspace{-8pt}
\end{center}
\end{table}

We select EfficientNet-B7 \cite{tan2019efficientnet} as our teacher model to guide our supernet training due to its state-of-the-art performance and relatively low computational cost comparing to ResNeXt-101 \cite{xie2017aggregated} and other manually designed models. We part the teacher model into 6 blocks by number of filters. The details of these blocks are presented in Table \ref{tab:design}.\\

\vspace{-10pt}
\noindent\textbf{Search space and supernet design.} We perform our search in two operation search spaces, both of which consist of variants of MobileNet V2's \cite{sandler2018mobilenetv2} Inverted Residual Block with Squeeze and Excitation \cite{hu2019squeeze}. We keep our first search space similar with most of the recent works \cite{tan2019mnasnet, tan2019efficientnet, chu2019scarletnas, chu2019moga} to facilitate fair comparison in Section \ref{sec:performance}. We search among convolution kernel sizes of \{3, 5, 7\} and expansion rates \{3, 6\}, six operations in total. For fast evaluation in Section \ref{sec:effctiveness} and \ref{sec:ablation}, a smaller search space with four operations (kernel sizes of \{3, 5\} and expansion rates \{3, 6\}) is used.

Upon operation search space, we further build a higher level search space to search for channel and layer numbers, as introduced in Section \ref{sect_scal}.
We search among three cells in each of the first 5 blocks and one in the last block.
The layer and channel numbers of each \emph{cell} is shown in Table \ref{tab:design}. The whole search space contains $2\times10^{17}$ models.\\

\vspace{-10pt}
\noindent\textbf{Training details} We separately train each \emph{cell} in the supernet for 20 epochs under the guidance of teacher's feature map in corresponding block. We use 0.002 as start learning rate for the first block and 0.005 for all the other blocks. We use Adam as our optimizer and reduce the learning rate by 0.9 every epoch. 

It takes 1 day to train a simple supernet (6 cells) using 8 NVIDIA GTX 2080Ti GPUs and 3 days for our extended supernet (16 cells).
With the help of Algorithm \ref{alg:eval}, our evaluation cost is about 0.6 GPU days.
To search for the best model under certain constraint, we perform Algorithm \ref{alg:search} on CPUs and the cost is less than one hour.

As for ImageNet retraining of searched models, we used the similar setting with \cite{tan2019efficientnet}: batchsize 4096, RMSprop optimizer with momentum 0.9 and initial learning rate of 0.256 which decays by 0.97 every 2.4 epochs.

\vspace{-5pt}
\subsection{Performance of searched models}\label{sec:performance}
\vspace{-6pt}

\begin{figure}[t!]
\vspace{-2pt}
\setlength{\abovecaptionskip}{0cm}
\setlength{\belowcaptionskip}{0.cm}
    \vspace{-12pt}
    \begin{subfigure}[t]{0.5\textwidth}
        \includegraphics[width=1\linewidth]{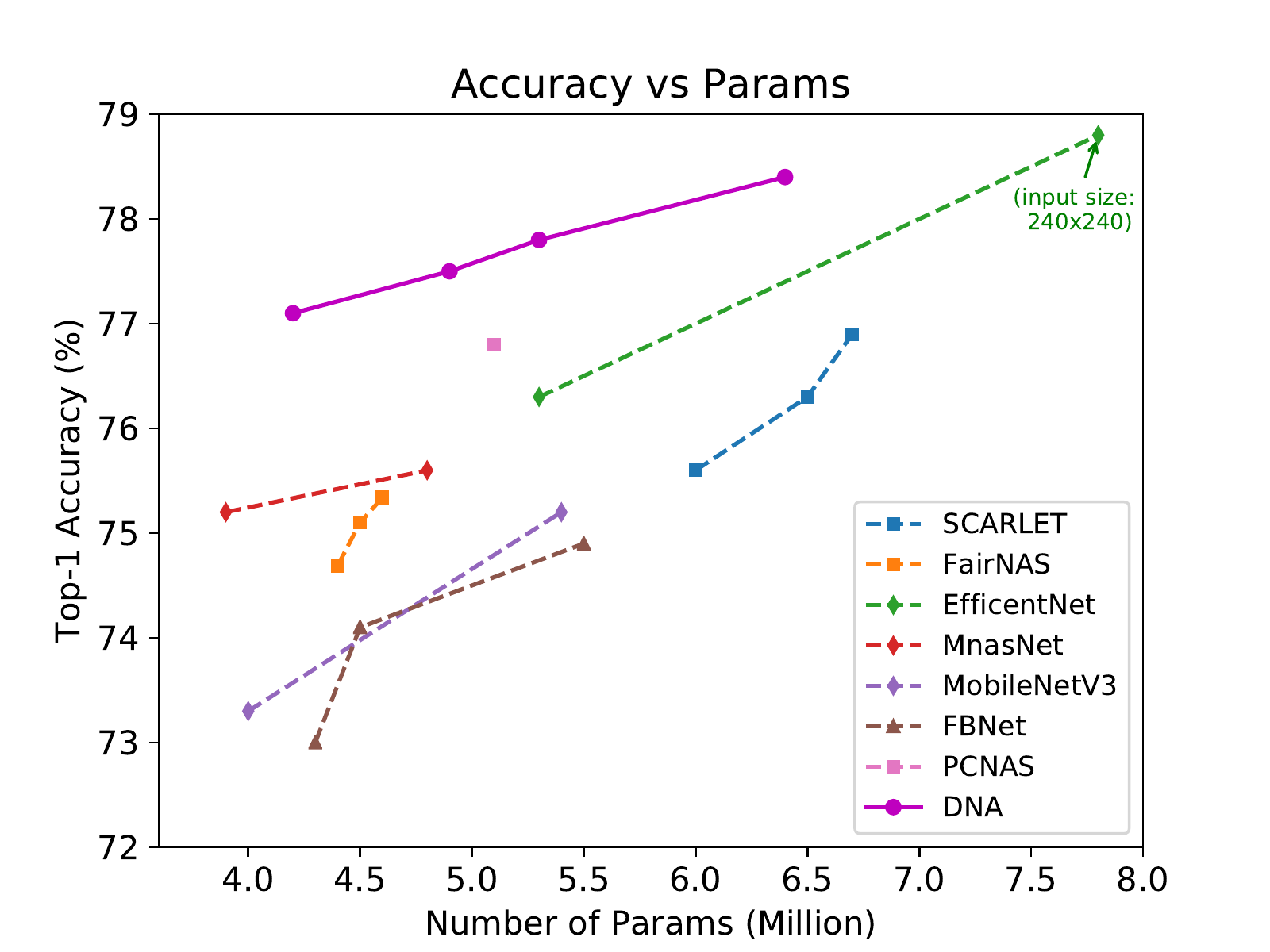}
    \end{subfigure}
    \begin{subfigure}[t]{0.5\textwidth}
        \includegraphics[width=1\linewidth]{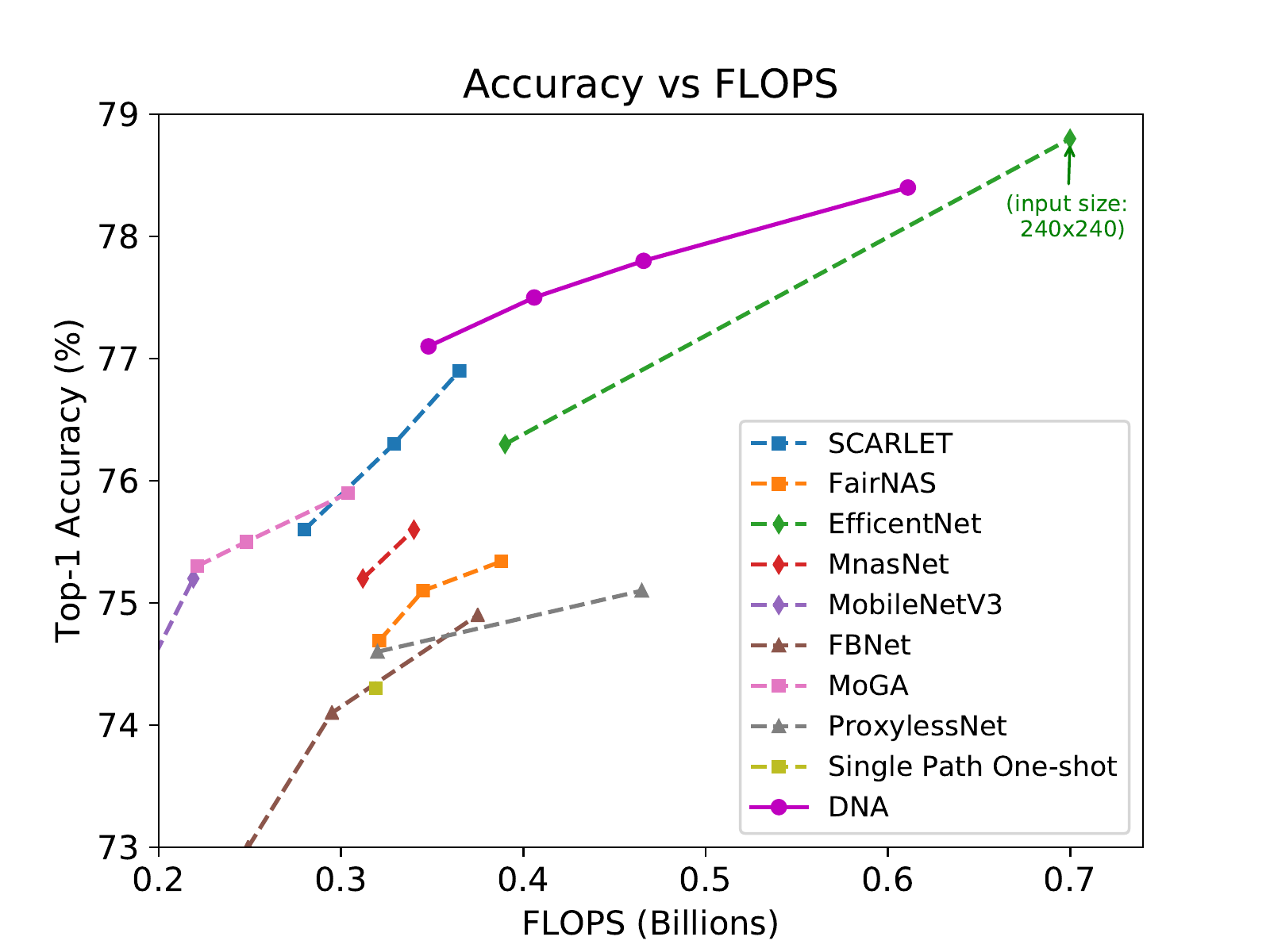}
    \end{subfigure}
    \caption{Trade-off of parameters-accuracy and FLOPs-accuracy on ImageNet.}
    \label{fig:paramsota}
\end{figure}

\begin{table}
\setlength{\abovecaptionskip}{0.cm}
\setlength{\belowcaptionskip}{0.cm}
\begin{center}
\caption{Comparison of state-of-the-art NAS models on ImageNet. The input size is $224\times 224$.}
\label{tab:imagenet}
\hskip-0.3cm
    \begin{tabular}{|p{2.9cm}|p{0.9cm}|p{0.9cm}|p{0.9cm}|p{0.9cm}|}
    \hline
    model                                           &Params        & FLOPs     & Acc@1     & Acc@5\\
    \hline\hline
    SPOS \cite{guo2019single}                      & -             & 319M      & 74.3\%    & -\\
    ProxylessNAS \cite{cai2019proxylessnas}        & 7.1M          & 465M      & 75.1\%    & 92.5\%\\
    FBNet-C \cite{wu2019fbnet}                     & -             & 375M      & 74.9\%    & -\\
    MobileNetV3 \cite{howard2019searching}         & 5.3M          & 219M      & 75.2\%    & -\\
    MnasNet-A3 \cite{tan2019mnasnet}               & 5.2M          & 403M      & 76.7\%    & 93.3\%\\
    FairNAS-A \cite{chu2019fairnas}                & 4.6M          & 388M      & 75.3\%    & 92.4\%\\
    MoGA-A \cite{chu2019moga}                      & 5.1M          & 304M      & 75.9\%    & 92.8\%\\
    SCARLET-A \cite{chu2019scarletnas}             & 6.7M          & 365M      & 76.9\%    & 93.4\%\\
    PC-NAS-S \cite{li2019improving}                & 5.1M          & -         & 76.8\%    & -\\
    MixNet-M \cite{tan2019mixconv}                 & 5.0M          & 360M      & 77.0\%    & 93.3\%\\
    EfficientNet-B0 \cite{tan2019efficientnet}     & 5.3M          & 399M      & 76.3\%    & 93.2\%\\
    \hline
    \hline
    DNA-a (ours)            & 4.2M          & 348M      & 77.1\%    & 93.3\%   \\
    DNA-b (ours)            & 4.9M          & 406M      & 77.5\%    & 93.3\%    \\
    DNA-c (ours)            & 5.3M          & 466M      & 77.8\%    & 93.7\%    \\
    DNA-d (ours)            & 6.4M          & 611M      & \textbf{78.4\%}   & \textbf{94.0\%}  \\
    \hline
    \end{tabular}
    \end{center}
\vspace{-8pt}
\end{table}

As shown in Table \ref{tab:imagenet}, our DNA models achieve the state-of-the-art results compared with the most recent NAS models. Searched under a FLOPs constraint of 350M, DNA-a surpasses SCARLET-A with 1.8M fewer parameters. For a fair comparison with EfficientNet-B0, DNA-b and DNA-c are obtained with target FLOPs of 399M and parameters of 5.3M respectively. Both of them outperform B0 by a large margin (1.2\% and 1.5\%). In particular, our DNA-d achieves 78.4\% top-1 accuracy with 6.4M parameters and 611M FLOPs. When tested with the same input size ($240\times240$) as EfficientNet-B1, DNA-d achieves 78.8\% top-1 accuracy, being evenly accurate but 1.4M smaller than B1. The accuracy of MixNet-M, who uses the more efficient MixConv operation that we don't use, is 0.4\% inferior to our smaller DNA-b.

Figure \ref{fig:paramsota} illustrates the curve of \textbf{Model size vs. Accuracy} and \textbf{FLOPs vs. Accuracy} for most recent NAS models. Our DNA models significantly mark a new state-of-the-art with much smaller model size and lower computation complexity.

To test the transfer ability of our model, We evaluate our model on two widely used transfer learning datasets, CIFAR-10 and CIFAR-100. Our models maintain superiority after the transfer. The result is shown in Table \ref{tab:cifar}.

\begin{table}
\setlength{\abovecaptionskip}{0.cm}
\setlength{\belowcaptionskip}{0.cm}
\begin{center}
\caption{Comparison of transfer learning performance of NAS models on CIFAR-10 and CIFAR-100. $^\dagger$: Our transfer learning results with officially released model. Accuracy within the parentheses are reported by the original paper.}
\label{tab:cifar}

\begin{tabular}{|l|l|l|}
\hline
Model & CIFAR-10 Acc & CIFAR-100 Acc\\
\hline\hline
MixNet-M\cite{tan2019mixconv} & 97.9\% & 87.4\% \\
EfficientNet-B0 & 98.0\%(98.1\%)$^\dagger$ & 87.1\%(88.1\%)$^\dagger$ \\
DNA-c (ours) & 98.3\% &88.3\% \\
\hline
\end{tabular}
\end{center}
\end{table}


\begin{figure}
\setlength{\abovecaptionskip}{-0cm}
\setlength{\belowcaptionskip}{0.1cm}
    \centering
    \includegraphics[width=1\linewidth]{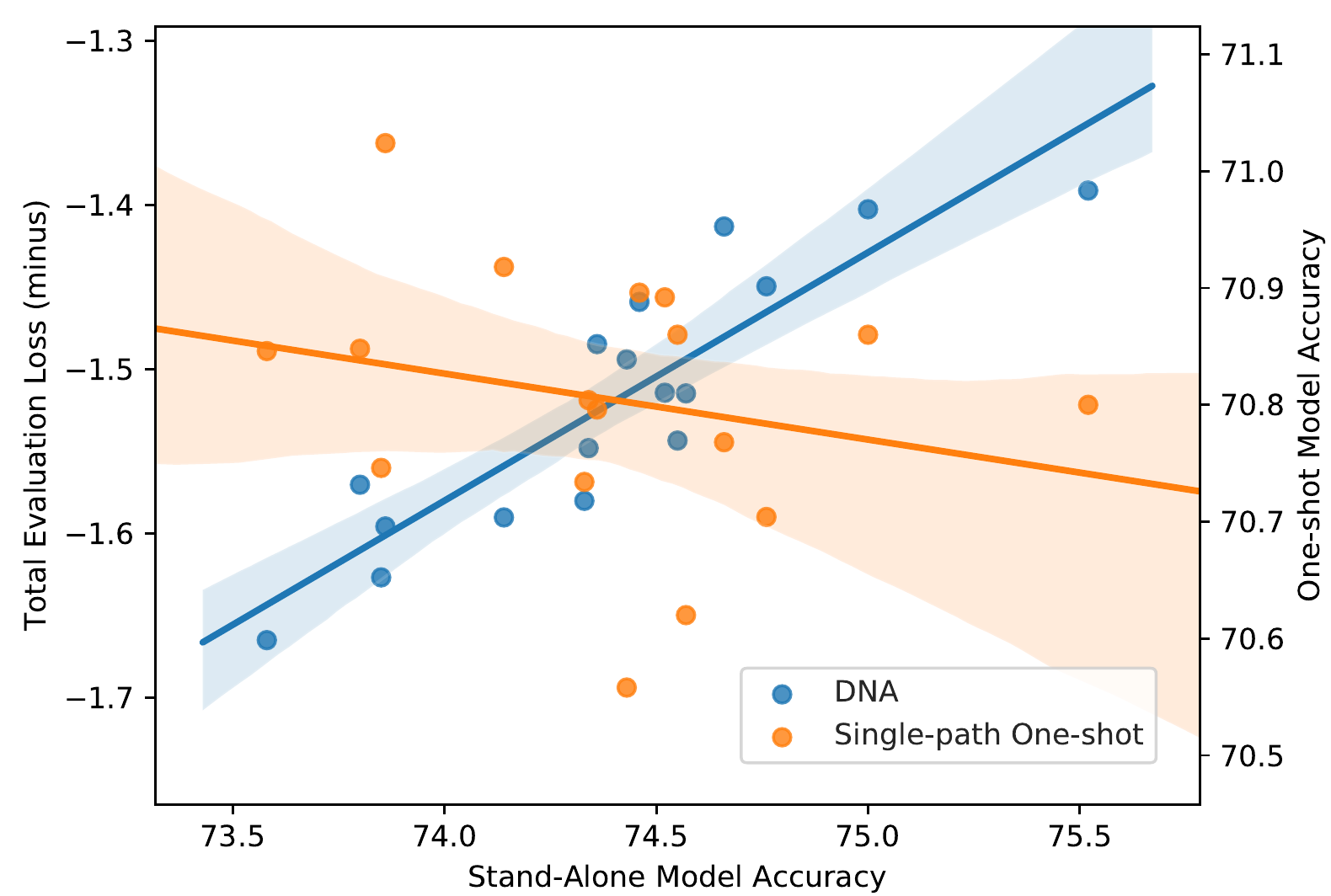}
    \caption{Comparison of ranking effectiveness for DNA and Single Path One-Shot\cite{guo2019single}}
    \label{fig:ranking}
\vspace{-4pt}
\end{figure}

\begin{figure*}[tp]
\setlength{\abovecaptionskip}{0cm}
\setlength{\belowcaptionskip}{0.1cm}
    \includegraphics[width=0.95\linewidth]{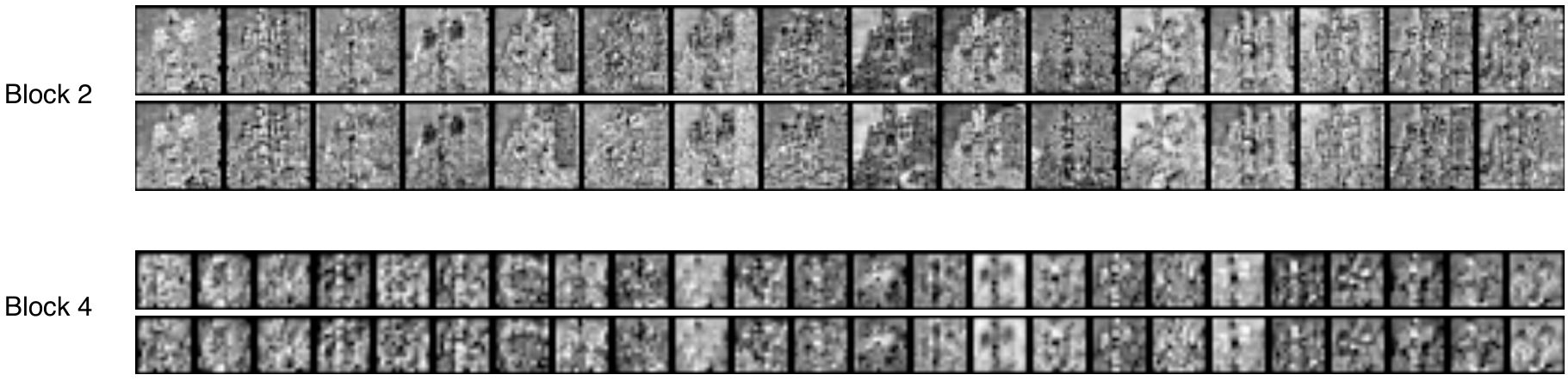}
    \caption{Feature map comparison between teacher (top) and student (bottom) of two blocks.}
    \label{fig:feature}
\end{figure*}

\begin{figure}
\vspace{-6pt}
\setlength{\abovecaptionskip}{0cm}
\setlength{\belowcaptionskip}{0.1cm}
    \centering
    \includegraphics[width=1\linewidth]{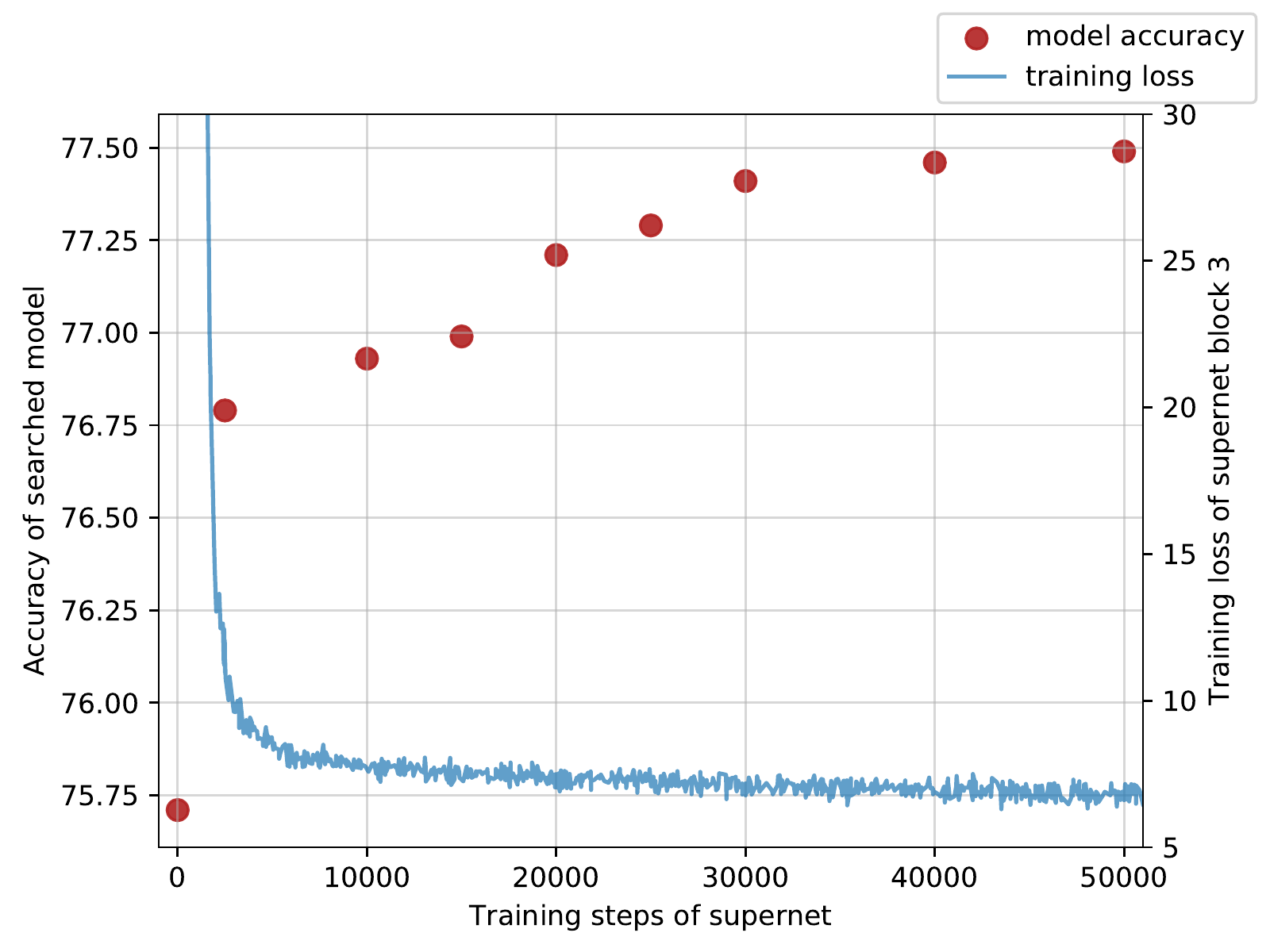}
    \caption{ImageNet accuracy of searched models and training loss of the supernet in training progress. }
    \label{fig:progress}
\end{figure}

\subsection{Effectiveness}\label{sec:effctiveness}
\vspace{-4pt}

\noindent\textbf{Model ranking.} To evaluate the effectiveness of our NAS method, we compared the model ranking abilities between our method and SPOS (Single Path One-shot\cite{guo2019single}) by visualizing the relationship between the evaluation metrics on proxy one-shot models and the actual accuracy of the stand-alone models. The two supernets are both 18 layers, with 4 candidate operations in each layer. The search space is described in Section \ref{sec:setup}.
We trained our supernet with 20 epochs for each block, adding up to 120 epochs in total. The supernet of Single Path One-shot is also trained for 120 epochs as they proposed\cite{guo2019single}. 

We sample 16 models from the search space and train them from scratch. As for model ranking test, we evaluate these sampled models in both supernets to get their predictive performance. The comparison of these two methods on model ranking is shown in Figure \ref{fig:ranking}. Each of the sampled model has two corresponding points in the figure,
representing the correlation between its predicted and true performance by two methods.
Figure \ref{fig:ranking} indicates that SPOS barely rank the candidate models correctly because the sub-nets are not fairly and fully trained as analyzed in Section \ref{defination}. While in our block-wise supernet, the predicted performance is highly correlated with the real accuracy of sampled models, which proves the effectiveness of our method.

\vspace{2pt}
\noindent\textbf{Training progress.} 
To analyse our supernet training process, we pick the intermediate models searched in every two training epochs (approximate to 5000 iterations) and retrain them to convergence. As shown in Figure \ref{fig:progress}, the accuracy of our searched models increase progressively as the training goes on until it converges between 16-th and 20-th epoch. It illustrates that the predictive metric of candidate models becomes more precise as the supernet converge. Note that the accuracy increase rapidly in the early stage with the same tendency of training loss decreasing, which evidences a correlation between accuracy of searched model and loss of supernet.

Part of the teacher and student feature map of block 2 and 4 at epoch 16 is shown in Figure \ref{fig:feature}. As we can see, our student supernet can imitate the the teacher extraordinarily well. The textures are extremely close at every channel, even on highly abstracted $14\times14$ feature maps. Which proves the effectiveness of our distillation training procedure. 
\vspace{-6pt}
\subsection{Ablation Study}\label{sec:ablation}
\vspace{-4pt}

\begin{table}
\setlength{\abovecaptionskip}{0.cm}
\setlength{\belowcaptionskip}{-0.cm}
\begin{center}
\caption{Impact of each component of DNA. Our strategy is better than S1 and S2. Adding cells to increase channel and layer variability can boost performance of searched model both with and without constraint.
}
\begin{tabular}{|p{1cm}|l|p{1.2cm}|p{0.9cm}|p{0.9cm}|p{0.9cm}|}
\hline
Strategy               &Cell      &Constrain      &Params     &Acc@1      &Acc@5\\
\hline
S1                  &                   &               &  5.18M    &77.0\%     &93.34\%\\
S2                  &                   &               &  5.58M    &77.15\%    &93.51\%\\
Ours                &                   &               &  5.69M    &77.49\%    &93.68\%\\
Ours                &\checkmark         &               &  6.26M    &77.84\%    &93.74\%\\
Ours                &                   &\checkmark     &  5.09M    &77.21\%	&93.50\%\\
Ours                &\checkmark         &\checkmark     &  5.28M    &77.38\%    &93.60\%\\
\hline
\end{tabular}
\label{tab:strategy}
\end{center}
\end{table}

\begin{table}
\begin{center}
\caption{Comparison of DNA with different teacher. Note that all the searched models are retrained from scratch without any supervision of the teacher. $^\dagger$:EfficientNet-B7 is tested with $224\times224$ input size, to be consistent with distillation procedure.}
\vspace{-5pt}
\begin{tabular}{|p{3.7cm}|p{0.9cm}|p{0.9cm}|p{0.9cm}|}
\hline
Model & Params & Acc@1 & Acc@5\\
\hline\hline
EfficientNet-B0 (Teacher) & 5.28M & 76.3\% & 93.2\% \\
\hline
DNA-B0 & 5.27M & 77.8\% & 93.7\% \\
\hline\hline
EfficientNet-B7 (Teacher) & 66M & 77.8\%$^\dagger$ & 93.8\%$^\dagger$ \\
\hline
DNA-B7 & 5.28M & 77.8\% & 93.7\% \\
\hline
DNA-B7-scale & 64.9M & 79.9\% & 94.9\% \\
\hline
\end{tabular}
\label{tab:selfdistill}
\end{center}
\vspace{-6pt}
\end{table}

\noindent\textbf{Distillation strategy.} We tested two progressive blockwise distillation strategy and compare their effectiveness with ours by experiments. All the three strategy is performed block by block by minimizing the MSE loss between feature maps of student supernet and the teacher. In strategy $S1$, the student is trained from scratch with all previous blocks in every stage. In strategy $S2$, the trained student parameters of the previous blocks is kept and freezed, thus those parameters are only used to generate the input feature map of current block. As discussed in Section \ref{sec:supernetdistill}, our strategy directly takes the teacher's previous feature map as input of the current block. The experimental results shown in Table \ref{tab:strategy} prove the superiority of our strategy.

\vspace{-2pt}
\noindent\textbf{Impact of multi-cell design.} To test the impact of multi-cell search, we preform DNA with single cell in each block for comparison. As shown in Table \ref{tab:strategy}, multi-cell search improves the top-1 accuracy of searched models by 0.2\% under the same constraint (5.3M) and 0.3\% for the best model in the search space without any constrain. Note that the single cell case of our method searched a model with lower parameters under the same constrain, this can be ascribed to the relatively lower variability of channel and layer numbers.

\noindent\textbf{Analysis of teacher-dependency.} To test the dependency of DNA on the performance of teacher model, EfficientNet-B0 is used as the teacher model to search for a student in the similar size. The results is shown in Table \ref{tab:selfdistill}. Surprisingly, performance of the model searched with EfficientNet-B0 is almost the same with the one searched with EfficientNet-B7, which means that the performance of our DNA method does not necessarily rely on high-performing teacher. Furthermore, the DNA-B0 outperforms its teacher by 1.5\% with the same model size, which proves that the performance of our architecture distillation is not restricted by the performance of the teacher. Thus, we can improve the structure of any model by self-distillation architecture search. Thirdly, DNA-B7 achieves same top-1 accuracy with its $12.5\times$ heavier teacher; by scaling our DNA-B7 to the similar model size as the supervising architecture, a more remarkable gain is further obtained. The scaled student outperforms its heavy teacher by 2.1\%, demonstrating the practicability and scalability of our DNA method.

\vspace{-8pt}
\section{Conclusion}
\vspace{-6pt}

In this paper, DNA, a novel architecture search method with block-wise supervision is proposed. We modularized the large search space into blocks to increase the effectiveness of one-shot NAS. We further designed a novel distillation approach to supervise the architecture search in a block-wise fashion. We then presented our multi-cell supernet design along with efficient evaluation and searching algorithms. We demonstrate that our searched architecture can surpass the teacher model and can achieve state-of-the-art accuracy on both ImageNet and two commonly used transfer learning datasets when trained from scratch without the helps of the teacher.
\vspace{-8pt}
\section*{Acknowledgements}
\vspace{-6pt}
We thank DarkMatter AI Research team for providing computational resources. C. Li and X. Chang gratefully acknowledge the support of Australian Research Council (ARC) Discovery Early Career Researcher Award (DECRA) under grant no. DE190100626, Air Force Research Laboratory and DARPA under agreement number FA8750-19-2-0501.

{\small
\bibliographystyle{ieee_fullname}
\bibliography{egbib}

\begin{thebibliography}{10}\itemsep=-1pt

\bibitem{AkimotoICML2019}
Youhei Akimoto, Shinichi Shirakawa, Nozomu Yoshinari, Kento Uchida, Shota
  Saito, and Kouhei Nishida.
\newblock Adaptive stochastic natural gradient method for one-shot neural
  architecture search.
\newblock In {\em Proceedings of the 36th International Conference on Machine
  Learning (ICML)}, pages 171--180, 2019.

\bibitem{ba2014deep}
Jimmy Ba and Rich Caruana.
\newblock Do deep nets really need to be deep?
\newblock In {\em Advances in Neural Information Processing Systems (NeurIPS)},
  pages 2654--2662, 2014.

\bibitem{baker2017designing}
Bowen Baker, Otkrist Gupta, Nikhil Naik, and Ramesh Raskar.
\newblock Designing neural network architectures using reinforcement learning.
\newblock {\em International Conference on Learning Representations (ICLR)},
  2017.

\bibitem{bender2018understanding}
Gabriel Bender, Pieter-Jan Kindermans, Barret Zoph, Vijay Vasudevan, and Quoc
  Le.
\newblock Understanding and simplifying one-shot architecture search.
\newblock In {\em Proceedings of the International Conference on Machine
  Learning (ICML)}, pages 550--559, 2018.

\bibitem{brock2018smash}
Andrew Brock, Theodore Lim, James~M Ritchie, and Nick Weston.
\newblock Smash: one-shot model architecture search through hypernetworks.
\newblock {\em International Conference on Learning Representations (ICLR)},
  2018.

\bibitem{cai2019proxylessnas}
Han Cai, Ligeng Zhu, and Song Han.
\newblock Proxylessnas: Direct neural architecture search on target task and
  hardware.
\newblock {\em International Conference on Learning Representations (ICLR)},
  2019.

\bibitem{chen2019renas}
Yukang Chen, Gaofeng Meng, Qian Zhang, Shiming Xiang, Chang Huang, Lisen Mu,
  and Xinggang Wang.
\newblock Renas: Reinforced evolutionary neural architecture search.
\newblock In {\em Proceedings of the IEEE Conference on Computer Vision and
  Pattern Recognition (CVPR)}, pages 4787--4796, 2019.

\bibitem{chu2019scarletnas}
Xiangxiang Chu, Bo Zhang, Jixiang Li, Qingyuan Li, and Ruijun Xu.
\newblock Scarletnas: Bridging the gap between scalability and fairness in
  neural architecture search.
\newblock {\em arXiv preprint arXiv:1908.06022}, 2019.

\bibitem{chu2019moga}
Xiangxiang Chu, Bo Zhang, and Ruijun Xu.
\newblock Moga: Searching beyond mobilenetv3.
\newblock {\em arXiv preprint arXiv:1908.01314}, 2019.

\bibitem{chu2019fairnas}
Xiangxiang Chu, Bo Zhang, Ruijun Xu, and Jixiang Li.
\newblock Fairnas: Rethinking evaluation fairness of weight sharing neural
  architecture search.
\newblock {\em arXiv preprint arXiv:1907.01845}, 2019.

\bibitem{devlin2018bert}
Jacob Devlin, Ming{-}Wei Chang, Kenton Lee, and Kristina Toutanova.
\newblock {BERT:} pre-training of deep bidirectional transformers for language
  understanding.
\newblock In {\em Proceedings of the 2019 Conference of the North American
  Chapter of the Association for Computational Linguistics: Human Language
  Technologies, {NAACL-HLT} 2019, Minneapolis, MN, USA, June 2-7, 2019, Volume
  1 (Long and Short Papers)}, pages 4171--4186, 2019.

\bibitem{dong2019searching}
Xuanyi Dong and Yi Yang.
\newblock Searching for a robust neural architecture in four gpu hours.
\newblock In {\em Proceedings of the IEEE Conference on Computer Vision and
  Pattern Recognition (CVPR)}, pages 1761--1770, 2019.

\bibitem{guo2019single}
Zichao Guo, Xiangyu Zhang, Haoyuan Mu, Wen Heng, Zechun Liu, Yichen Wei, and
  Jian Sun.
\newblock Single path one-shot neural architecture search with uniform
  sampling.
\newblock {\em arXiv preprint arXiv:1904.00420}, 2019.

\bibitem{hinton2014distilling}
Geoffrey Hinton, Oriol Vinyals, and Jeff Dean.
\newblock Distilling the knowledge in a neural network.
\newblock In {\em NIPS Deep Learning Workshop}, 2014.

\bibitem{howard2019searching}
Andrew Howard, Mark Sandler, Grace Chu, Liang-Chieh Chen, Bo Chen, Mingxing
  Tan, Weijun Wang, Yukun Zhu, Ruoming Pang, Vijay Vasudevan, et~al.
\newblock Searching for mobilenetv3.
\newblock In {\em Proceedings of the IEEE International Conference on Computer
  Vision (ICCV)}, pages 1314--1324, 2019.

\bibitem{hu2019squeeze}
J Hu, L Shen, S Albanie, G Sun, and E Wu.
\newblock Squeeze-and-excitation networks.
\newblock {\em IEEE transactions on pattern analysis and machine intelligence},
  2019.

\bibitem{li2019improving}
Xiang Li, Chen Lin, Chuming Li, Ming Sun, Wei Wu, Junjie Yan, and Wanli Ouyang.
\newblock Improving one-shot nas by suppressing the posterior fading.
\newblock {\em arXiv preprint arXiv:1910.02543}, 2019.

\bibitem{liang2020computation}
Feng Liang, Chen Lin, Ronghao Guo, Ming Sun, Wei Wu, Junjie Yan, and Wanli
  Ouyang.
\newblock Computation reallocation for object detection.
\newblock {\em International Conference on Learning Representations (ICLR)},
  2020.

\bibitem{liu2019darts}
Hanxiao Liu, Karen Simonyan, and Yiming Yang.
\newblock Darts: Differentiable architecture search.
\newblock {\em International Conference on Learning Representations (ICLR)},
  2019.

\bibitem{negrinho2018deeparchitect}
Renato Negrinho and Geoff Gordon.
\newblock Deeparchitect: Automatically designing and training deep
  architectures.
\newblock {\em International Conference on Learning Representations (ICLR)},
  2018.

\bibitem{passalis2018learning}
Nikolaos Passalis and Anastasios Tefas.
\newblock Learning deep representations with probabilistic knowledge transfer.
\newblock In {\em Proceedings of the European Conference on Computer Vision
  (ECCV)}, pages 268--284, 2018.

\bibitem{romero2015fitnets}
Adriana Romero, Nicolas Ballas, Samira~Ebrahimi Kahou, Antoine Chassang, Carlo
  Gatta, and Yoshua Bengio.
\newblock Fitnets: Hints for thin deep nets.
\newblock {\em International Conference on Learning Representations (ICLR)},
  2015.

\bibitem{sandler2018mobilenetv2}
Mark Sandler, Andrew Howard, Menglong Zhu, Andrey Zhmoginov, and Liang-Chieh
  Chen.
\newblock Mobilenetv2: Inverted residuals and linear bottlenecks.
\newblock In {\em Proceedings of the IEEE Conference on Computer Vision and
  Pattern Recognition (CVPR)}, pages 4510--4520, 2018.

\bibitem{sciuto2019evaluating}
Christian Sciuto, Kaicheng Yu, Martin Jaggi, Claudiu Musat, and Mathieu
  Salzmann.
\newblock Evaluating the search phase of neural architecture search.
\newblock {\em International Conference on Learning Representations (ICLR)},
  2020.

\bibitem{shipp1985segregation}
Stewart Shipp and Semir Zeki.
\newblock Segregation of pathways leading from area v2 to areas v4 and v5 of
  macaque monkey visual cortex.
\newblock {\em Nature}, 315(6017):322--324, 1985.

\bibitem{tan2019mnasnet}
Mingxing Tan, Bo Chen, Ruoming Pang, Vijay Vasudevan, Mark Sandler, Andrew
  Howard, and Quoc~V Le.
\newblock Mnasnet: Platform-aware neural architecture search for mobile.
\newblock In {\em Proceedings of the IEEE Conference on Computer Vision and
  Pattern Recognition (CVPR)}, pages 2820--2828, 2019.

\bibitem{tan2019efficientnet}
Mingxing Tan and Quoc Le.
\newblock Efficientnet: Rethinking model scaling for convolutional neural
  networks.
\newblock In {\em International Conference on Machine Learning (ICML)}, pages
  6105--6114, 2019.

\bibitem{tan2019mixconv}
Mingxing Tan and Quoc~V Le.
\newblock Mixconv: Mixed depthwise convolutional kernels.
\newblock In {\em Proceedings of the 30th British Machine Vision Conference
  (BMVC)}, 2019.

\bibitem{vaswani2017attention}
Ashish Vaswani, Noam Shazeer, Niki Parmar, Jakob Uszkoreit, Llion Jones,
  Aidan~N Gomez, {\L}ukasz Kaiser, and Illia Polosukhin.
\newblock Attention is all you need.
\newblock In {\em Advances in Neural Information Processing Systems (NeurIPS)},
  pages 5998--6008, 2017.

\bibitem{wang2018progressive}
Hui Wang, Hanbin Zhao, Xi Li, and Xu Tan.
\newblock Progressive blockwise knowledge distillation for neural network
  acceleration.
\newblock In {\em Proceedings of the International Joint Conference on
  Artificial Intelligence (IJCAI)}, pages 2769--2775, 2018.

\bibitem{wu2019fbnet}
Bichen Wu, Xiaoliang Dai, Peizhao Zhang, Yanghan Wang, Fei Sun, Yiming Wu,
  Yuandong Tian, Peter Vajda, Yangqing Jia, and Kurt Keutzer.
\newblock Fbnet: Hardware-aware efficient convnet design via differentiable
  neural architecture search.
\newblock In {\em Proceedings of the IEEE Conference on Computer Vision and
  Pattern Recognition (CVPR)}, pages 10734--10742, 2019.

\bibitem{xie2017aggregated}
Saining Xie, Ross Girshick, Piotr Doll{\'a}r, Zhuowen Tu, and Kaiming He.
\newblock Aggregated residual transformations for deep neural networks.
\newblock In {\em Proceedings of the IEEE Conference on Computer Vision and
  Pattern Recognition (CVPR)}, pages 1492--1500, 2017.

\bibitem{yang2019evaluation}
Antoine Yang, Pedro~M Esperan{\c{c}}a, and Fabio~M Carlucci.
\newblock Nas evaluation is frustratingly hard.
\newblock {\em International Conference on Learning Representations (ICLR)},
  2020.

\bibitem{yim2017gift}
Junho Yim, Donggyu Joo, Jihoon Bae, and Junmo Kim.
\newblock A gift from knowledge distillation: Fast optimization, network
  minimization and transfer learning.
\newblock In {\em Proceedings of the IEEE Conference on Computer Vision and
  Pattern Recognition (CVPR)}, pages 4133--4141, 2017.

\bibitem{zhang2017knowledge}
Zhi Zhang, Guanghan Ning, and Zhihai He.
\newblock Knowledge projection for deep neural networks.
\newblock {\em arXiv preprint arXiv:1710.09505}, 2017.

\bibitem{zhong2018practical}
Zhao Zhong, Junjie Yan, Wei Wu, Jing Shao, and Cheng-Lin Liu.
\newblock Practical block-wise neural network architecture generation.
\newblock In {\em Proceedings of the IEEE Conference on Computer Vision and
  Pattern Recognition (CVPR)}, pages 2423--2432, 2018.

\bibitem{zoph2017neural}
Barret Zoph and Quoc~V Le.
\newblock Neural architecture search with reinforcement learning.
\newblock {\em International Conference on Learning Representations (ICLR)},
  2017.

\end{thebibliography}
}
\clearpage
\appendix
\renewcommand{\appendixname}{Appendix~\Alph{section}}
\section{Appendix}%
    \subsection{Model Architectures}%
    \begin{figure*}[hb]
        \begin{center}
        \includegraphics[width=0.9\linewidth]{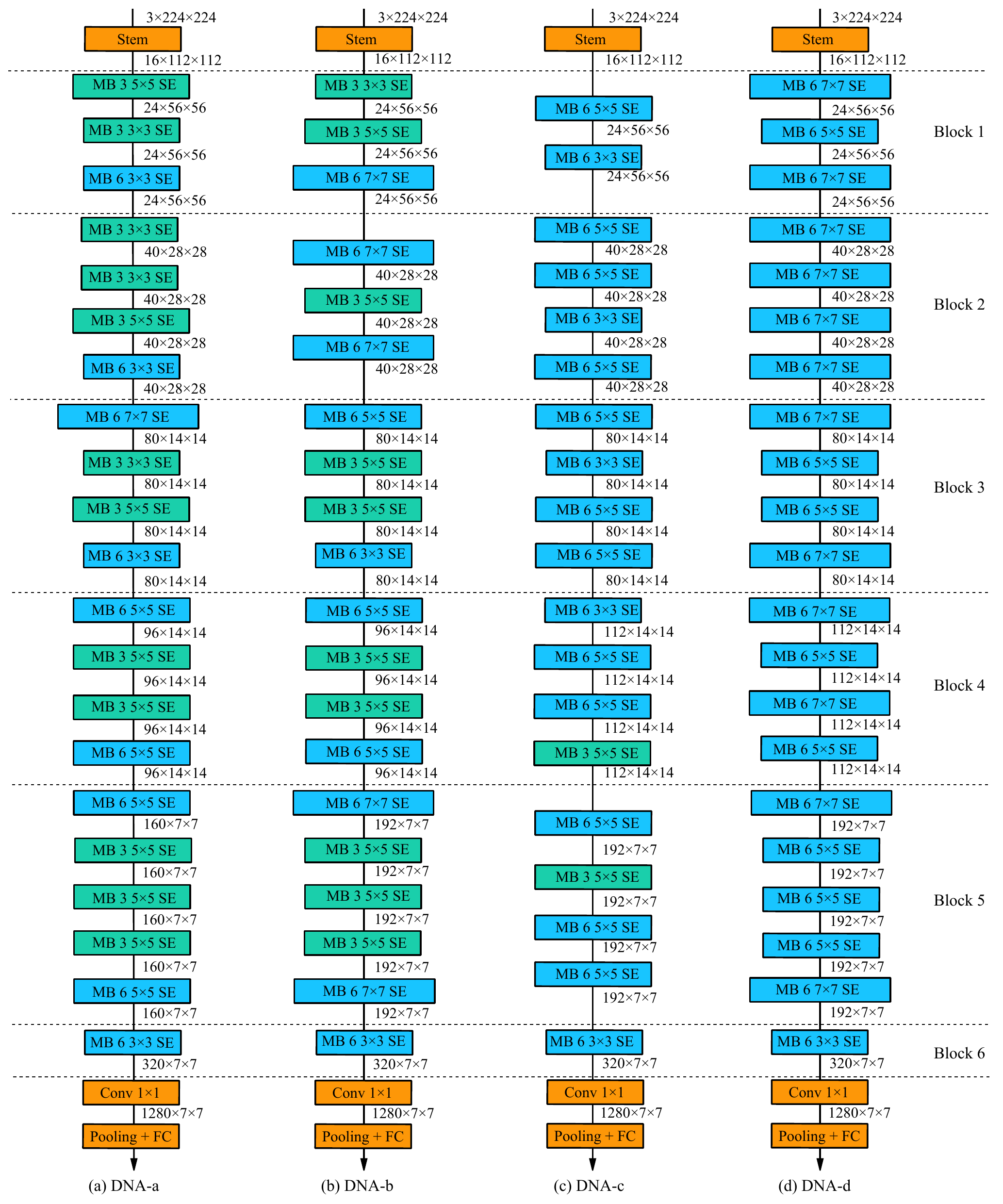}
        \caption{Architectures of DNA-a,b,c,d. `MB $x$ $y\times y$' stands for an Inverted bottleneck convolution module with expand rate $x$ and kernel size $y$.}
        \label{fig:arch_detail}
        \end{center}
    \end{figure*}
\end{document}